\documentclass[numbers]{article}

\usepackage[main, final]{neurips_2026}

\usepackage[utf8]{inputenc}
\usepackage[T1]{fontenc}
\usepackage{hyperref}
\usepackage{url}
\usepackage{booktabs}
\usepackage{amsfonts}
\usepackage{amsmath,amssymb,amsthm,mathtools}
\usepackage{nicefrac}
\usepackage{microtype}
\usepackage{xcolor}
\usepackage{graphicx}
\graphicspath{{figures/}}
\usepackage{enumitem}
\usepackage{multirow}
\usepackage{array}
\usepackage{algorithm}
\usepackage{algpseudocode}
\usepackage{tikz}
\usetikzlibrary{positioning,fit,arrows.meta,calc,shapes.geometric,shapes.misc,backgrounds}

\newtheorem{definition}{Definition}
\newtheorem{assumption}{Assumption}
\newtheorem{proposition}{Proposition}
\newtheorem{theorem}{Theorem}

\newcommand{\method}{FN-HFL}
\newcommand{\sat}{\mathcal{V}}
\newcommand{\edges}{\mathcal{E}}
\newcommand{\graph}{\mathcal{G}}
\newcommand{\weights}{\mathbf{W}}

\newcommand{\swcap}{\mathbf{c}}
\newcommand{\contact}{\mathcal{C}}
\newcommand{\agg}{\operatorname{Agg}}

\title{FractalNet-Based Heterogeneous Federated Learning for Orbital Edge Intelligence in Satellite Mega-Constellations: A Wildfire Case Study}

\author{
Sai Puppala \\
School of Computing\\
Southern Illinois University\\
Carbondale, IL 62901, USA \\
Email: \texttt{sai.puppala@siu.edu} \\
\And
Bhabani P. Sinha \\
Computer Science Engineering\\
Sister Nivedita University\\
Kolkata, India\\
Email: \texttt{bhabani.ps@snuniv.ac.in}\\
\And
Koushik Sinha\thanks{Corresponding author}  \\
School of Computing \\
Southern Illinois University \\
Carbondale, IL 62901, USA \\
Email: \texttt{koushik.sinha@siu.edu} \\
}

\begin{document}
\maketitle

\begin{abstract}
Satellite mega-constellations are emerging as large-scale sensing, communication, and computation fabrics, yet their learning architectures remain largely inherited from terrestrial federated learning and ground-centric mission operations--- ill-suited to satellites that differ by orders of magnitude in Size, Weight, Power, and Cost (SWAP-C), radiation tolerance, link availability, and propagation delay. We propose a heterogeneous federated learning method based on the FractalNet architecture for orbital edge intelligence. We formalize contact-window-constrained, depth-heterogeneous federated optimization and introduce a distributed path scheduler that assigns model depth as a function of SWAP-C constraints, predicted inter-satellite contacts, and training statistics. To reduce message overhead and energy consumption, each tier pools updates periodically rather than at every contact opportunity, and a three-tier agentic control plane governs in-space scheduling, anomaly escalation, and policy-governed autonomy. As a case study, we apply the framework to wildfire detection, where each orbital shell naturally learns a different semantic level of situational awareness: pixel-scale thermal anomalies at low Earth orbit (LEO), regional fire-front dynamics at medium Earth orbit (MEO), and larger-scale risk propagation at geostationary or high Earth orbit (GEO/HEO). Experiments on simulated mega-constellations validate the approach across convergence, communication efficiency, energy adaptation, scheduled-pooling savings, robustness, and latency.
\end{abstract}

\section{Introduction}

Mega-constellations shift the unit of space-system design from an individually controlled spacecraft to a massive, time-varying distributed system. Thousands of satellites observe, route, infer, and coordinate under constraints that differ sharply from terrestrial cloud and mobile-edge settings: inter-satellite links (ISLs) appear and disappear according to orbital mechanics; radiation induces non-adversarial but security-relevant compute and memory faults; downlink opportunities are intermittent; and onboard compute spans from radiation-hardened microprocessors to accelerator-class edge devices.

Federated learning (FL) appears natural for this setting because raw telemetry, imagery, and RF observations remain local while model updates are exchanged. However, standard FL carries an implicit homogeneity assumption: every client trains the same model architecture, possibly at different speeds. In satellite constellations this assumption is structurally wrong. A weak LEO satellite may only support a shallow feature extractor, whereas a high-compute GEO/HEO node may support full sequence modeling and robust aggregation. Designing for the weakest node wastes high-compute satellites; designing for the strongest node excludes the majority of the constellation.

This paper proposes \method{}, a co-designed architecture in which \emph{the topology is part of the learning algorithm}. Rather than layering heterogeneous FL on top of an arbitrary communication substrate, \method{} maps FractalNet-style multi-path structure to model-depth allocation. Short paths correspond to shallow-prefix training, medium paths to partial-depth training, and deep paths to full-model training. A scheduler assigns satellite roles and path depths using current SWaP-C state, forecast contact windows, model-layer statistics, and trust signals. Agents onboard satellites operate at three timescales: micro-agents for local reflexes, meso-agents for regional aggregation, and macro-agents for constellation-level planning.

To illustrate how naturally this architecture maps onto real operational scenarios, we develop a wildfire detection use case in Section~\ref{sec:wildfire} that shows how each tier of the constellation learns semantically distinct information aligned with its orbit, revisit rate, and computational capability. This grounding motivates the formal architecture that follows.

\paragraph{Contributions.} We make five contributions.
\begin{enumerate}[leftmargin=*,nosep]
    \item We introduce \method{}, a topology-native heterogeneous FL architecture for satellite mega-constellations that ties FractalNet path length to model depth and aggregation responsibility.
    \item We formalize depth-heterogeneous federated optimization over scheduled orbital contact graphs, including layer-wise participation masks, staleness bounds, and SWaP-C constraints.
    \item We present a distributed scheduler that selects shallow, medium, and deep training paths and defines how LEO, MEO, and GEO/HEO nodes exchange updates without downlinking raw data.
    \item We define a depth-stratified aggregation and model-consistency protocol that synchronizes different model prefixes across heterogeneous satellites.
    \item We introduce \emph{scheduled pooling}, a tier-specific update-aggregation cadence (30 minutes at LEO, 6 hours at MEO, 24 hours at GEO/HEO) that decouples transmission frequency from raw contact-window availability, substantially reducing per-node energy expenditure relative to opportunistic, every-contact transmission.
    \item We provide a NeurIPS-style experimental protocol with simulator assumptions, baselines, metrics, and ablations, anchored by a concrete wildfire detection scenario that validates the depth-hierarchy design.
\end{enumerate}

\section{Motivating Scenario: Wildfire Detection Across a Mega-Constellation}
\label{sec:wildfire}

To make the \method{} architecture tangible before introducing formalism, we trace a wildfire breaking out in a remote forested region. This scenario is not merely illustrative. It exposes exactly the structural properties---multi-scale sensing, information aggregation hierarchy, compute heterogeneity, and intermittent connectivity---that motivated the design. Crucially, the information that each tier of the constellation can usefully contribute maps directly onto the depth of model that tier can afford to train.

\subsection{Shell 3 --- LEO sensing nodes: raw signal detection}

The first responders are the large population of low-altitude LEO nodes, passing over any given region every few minutes at typical altitudes between 500 and 600 km. These are the most constrained devices in the constellation: limited battery reserves, modest processors, and narrow contact windows measured in minutes per pass. But they carry multispectral and thermal sensors, and their low altitude gives them the spatial resolution needed to detect early-stage fire signatures.

Specifically, LEO nodes are positioned to observe: anomalous temperature clusters in the thermal infrared band (LWIR), suggesting surface heating above ambient; localized shifts in near-infrared (NIR) reflectance consistent with vegetation beginning to combust; aerosol density spikes in shortwave infrared (SWIR) channels indicating early smoke column formation; and rapid frame-to-frame pixel changes in registered image sequences consistent with active combustion rather than agricultural activity or industrial heat sources.

Each LEO node trains only the \emph{shallow prefix} of the shared model---the early convolutional or patch-embedding layers that learn to recognize these low-level, pixel-scale features from raw sensor streams. No single LEO node sees the whole fire. Each sees a tile, a moment, a local anomaly. Their gradient updates are cheap to compute, covering only the early model layers, and are relayed during brief contact windows to the nearest MEO regional aggregator. The shallow updates from dozens of LEO passes, each observing a different tile of the affected region, collectively train a lower-layer feature extractor that generalizes across fire intensities, vegetation types, smoke opacity levels, and sensor viewing angles---precisely because the constellation's broad spatial coverage provides the diversity that no single ground sensor or individual satellite pass could achieve.

\subsection{Shell 2 --- MEO regional nodes: contextual pattern recognition}

MEO satellites orbit higher and move more slowly relative to the ground, giving them longer contact windows with LEO nodes below and global nodes above, and a wider instantaneous field of view. They receive shallow updates from dozens of LEO passes and aggregate them into a regional picture. But they also train their own \emph{medium-depth layers} on this aggregated signal---layers that learn to recognize spatial propagation patterns and regional context that no individual LEO pass can see.

The medium layers trained at MEO learn: the characteristic elongated shape of a wind-driven fire front, distinguishable from point-source industrial fires; the correlation between temperature anomaly clusters and terrain features such as ridge lines, drainage channels, and fuel-break boundaries that shape fire propagation; the temporal cadence signature of a spreading fire versus a contained or dying one, visible only when comparing updates from multiple LEO passes over the same area; and the statistical relationship between early SWIR aerosol signals and subsequent thermal anomaly patterns, enabling forward prediction of fire spread direction.

A single MEO node might coordinate updates from LEO satellites spanning several hundred kilometers of fire perimeter, synthesizing what looked like scattered thermal noise at the LEO level into a coherent, moving boundary. This is precisely where the medium-depth model earns its architectural value: the early layers, trained by LEO nodes, learn what fire looks like locally; the middle layers, trained and aggregated at MEO, learn how fires \emph{behave} regionally. The MEO node does not need the raw pixels. It receives compressed gradient updates from LEO nodes, aggregates them robustly across the shallow layers, extends the model forward through the medium layers using its own regional training data, and forwards a richer update upward. The communication savings are substantial: the MEO aggregator transmits a single regional summary rather than relaying dozens of raw LEO update packets to the global tier.

\subsection{Shell 1 --- GEO/HEO global nodes: multi-region situational awareness}

High-compute global nodes see across entire continents. They receive regional summaries from multiple MEO aggregators simultaneously---one covering a Pacific coast range, another tracking an interior mountain fire complex, a third detecting a secondary ignition hundreds of kilometers from the original event driven by long-range ember transport. The \emph{deep layers} of the model, trained and maintained exclusively at this tier, learn cross-regional and multi-event patterns that no lower tier could represent.

Specifically, the deep layers learn: the statistical relationship between current large-scale atmospheric patterns (wind fields, humidity gradients, pressure systems) and fire propagation severity across multiple simultaneous events; historical correlations between early fire signatures in a given region and eventual burned area, enabling rapid severity classification that informs resource pre-positioning; the characteristic temporal signature of ember-driven spotting events, which create secondary ignitions kilometers ahead of the main front and which only become visible when comparing observations across widely separated MEO regions; and multi-year climate-driven risk signatures that allow the model to flag regions currently exhibiting pre-fire conditions even before any thermal anomaly is detected.

These are patterns that no LEO or MEO node could learn in isolation. They require integrating signals that are spatially separated by hundreds to thousands of kilometers and temporally separated by hours to days. The GEO/HEO tier provides both the computational resources to train these deep representations and the persistent wide-area view that makes the training signal meaningful.

\subsection{Why this maps cleanly onto \method{}}

The wildfire scenario is not a post-hoc rationalization of the architecture. It reveals the principled reason why depth should track orbital hierarchy rather than being assigned arbitrarily. The information content available at each tier is genuinely hierarchical: raw sensor features at LEO, spatial propagation context at MEO, multi-event systemic patterns at GEO/HEO. A model architecture that assigns the same depth to all tiers either wastes the compute of high-orbit nodes on shallow features they could learn trivially or asks constrained LEO nodes to train representations that require multi-region context they cannot observe.

\method{} makes the tier-depth correspondence explicit and schedulable. When a LEO satellite's battery is degraded, the scheduler reduces its depth assignment; the shallow feature extractor still benefits from its local observations, and no MEO or GEO layer is starved of contributors. When a MEO node loses an ISL contact window, its regional update is marked stale and down-weighted in the global aggregation; the fire-front propagation layers still receive contributions from other MEO nodes covering adjacent regions. When a GEO node detects an anomalous gradient in a deep layer, its trust score is reduced and the aggregation rule falls back to a more conservative cross-node consensus, protecting the multi-event situational awareness layers that downstream emergency response systems depend on.

The wildfire scenario also exposes the novel security risk \method{} must address. A compromised deep node---one that has received a malicious firmware update or is operating under a sophisticated adversarial attack---can inject a gradient into the lower layers that all LEO nodes use for basic thermal detection, while simultaneously reporting plausible fire-front propagation signals at the medium layers. Shallow LEO nodes, which never inspect the deep layers, cannot detect this attack. The defense must therefore operate layer-wise, cross-referencing gradient statistics across nodes at each depth level, using the large population of LEO-trained shallow updates as a reference distribution against which to test contributions from deeper nodes to the shared lower layers.

\section{Related Work}

\paragraph{Fractal architectures and anytime computation.}
FractalNet introduced self-similar neural macro-architectures with paths of different lengths and drop-path regularization, showing that shallow and deep subnetworks can be jointly trained and later extracted as useful fixed-depth subnetworks~\citep{larsson2016fractalnet,larsson2017fractalnet}. \method{} reinterprets this property at system scale: instead of randomly dropping paths on one machine, orbital contact constraints and node resources determine which path depths participate in each round.

\paragraph{Heterogeneous federated learning.}
HeteroFL, FjORD, and DepthFL study FL with heterogeneous local model sizes or depths~\citep{diao2021heterofl,horvath2021fjord,kim2023depthfl}. These methods motivate model heterogeneity but do not co-design client capability with a hierarchical orbital topology. \method{} differs by making communication path length the mechanism for assigning model depth and aggregation responsibility---a co-design that the wildfire scenario shows is semantically motivated, not merely an engineering compromise.

\paragraph{Satellite federated learning.}
Prior satellite FL systems study hierarchical aggregation, decentralized LEO training, offloading, and straggler mitigation under dynamic links~\citep{razmi2022ground,zhai2024fedleo,dfedsat2024,hisatfl2025}. These works establish feasibility, but most retain homogeneous model assumptions or treat topology as a routing constraint rather than a learning primitive.

\paragraph{Robust aggregation and security.}
Byzantine-robust FL aggregation rules such as Krum, coordinate-wise median, and trimmed mean protect against malicious client updates~\citep{blanchard2017krum,yin2018byzantine}. In \method{}, adversarial and radiation-induced corruptions may be depth-targeted, as the wildfire security discussion illustrates. This motivates depth-stratified aggregation that cannot be reduced to existing layer-agnostic robust rules.

\section{System Model}

\subsection{Orbital graph and hierarchy}

Let \(\graph_t=(\sat,\edges_t)\) be the time-indexed contact graph at training round \(t\), where vertices are satellites and edges are active ISLs. Each node \(i\in\sat\) has a shell label \(s_i\in\{1,2,3\}\), with Shell 3 denoting constrained low-altitude sensing nodes, Shell 2 denoting regional aggregation nodes, and Shell 1 denoting high-compute global aggregation nodes. The contact schedule \(\contact=\{\edges_t\}_{t=1}^T\) is known over a finite horizon from orbital propagation, but execution may deviate because of link outages, attitude maneuvers, or space-weather events.

\begin{figure}[t]
    \centering
    \includegraphics[width=\linewidth]{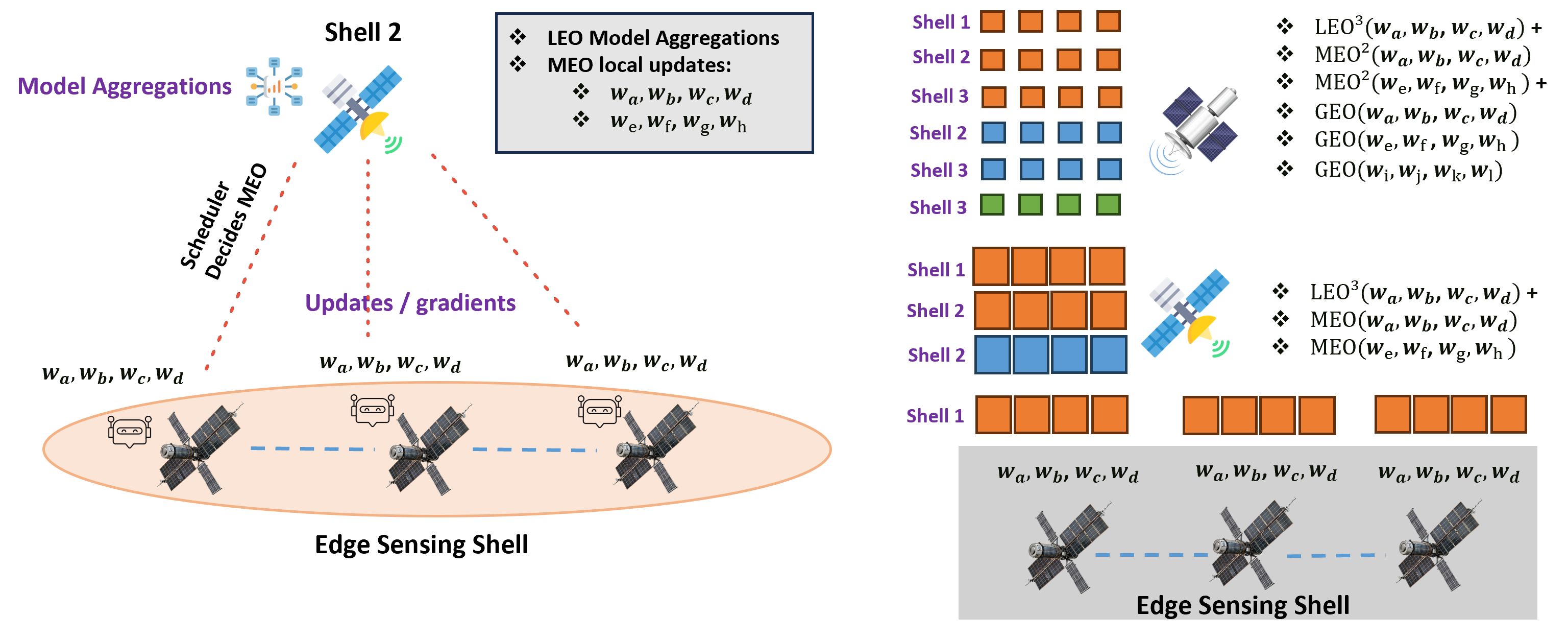}
    \caption{\textbf{FN-HFL architecture for depth-heterogeneous satellite federated learning.}
    Edge sensing satellites in Shell~3 train only shallow model blocks, producing local updates such as \(w_a,w_b,w_c,w_d\) from onboard observations. These shallow updates are transmitted as gradients to regional Shell~2 satellites, where MEO nodes perform LEO model aggregation and train additional medium-depth blocks, e.g., \((w_a,\ldots,w_h)\). Shell~1 high-compute GEO/HEO nodes receive regional summaries and aggregate full-depth updates, including deeper blocks \((w_i,\ldots,w_l)\), to maintain the global model. The right side illustrates the FractalNet-native depth hierarchy: shorter paths correspond to shallow LEO training, intermediate paths to MEO regional learning, and longer paths to full-depth GEO/HEO aggregation. This topology-depth mapping enables resource-aware learning in which constrained satellites contribute useful shallow features while higher-orbit nodes learn regional and global wildfire semantics without requiring raw data downlink.}
\label{fig:fnhfl_architecture}
\end{figure}

Each node has a time-varying resource vector
\begin{equation}
\swcap_i(t)=\big(E_i(t),P_i(t),M_i(t),B_i(t),R_i(t),Q_i(t)\big),
\end{equation}
where \(E\) is energy budget, \(P\) processor availability, \(M\) memory, \(B\) link bandwidth, \(R\) radiation/fault-risk state, and \(Q\) queue pressure. We write \(\mathcal{N}_i(t)\) for active neighbors of \(i\).

\subsection{Depth-heterogeneous model}

The global model is decomposed into \(L\) ordered blocks,
\begin{equation}
    f(x;\weights)=f_L\circ f_{L-1}\circ\cdots\circ f_1(x), \qquad \weights=\{W_1,\ldots,W_L\}.
\end{equation}
Each satellite trains a prefix or segment determined by a depth assignment \(d_i(t)\in\{1,\ldots,L\}\). For the canonical three-tier system, shallow LEO nodes train \(d_S\) layers, medium MEO nodes train \(d_M\) layers, and deep GEO/HEO nodes train all \(L\) layers:
\begin{equation}
    d_i(t) \in \{d_S,d_M,L\}, \qquad 1\leq d_S < d_M < L.
\end{equation}
In the wildfire scenario and throughout the experimental section, we use \(L=10\), \(d_S=3\), and \(d_M=6\): LEO nodes contribute gradients for \(\{W_1,W_2,W_3\}\) (pixel-scale feature extraction), MEO nodes for \(\{W_1,\ldots,W_6\}\) (regional propagation modeling), and GEO/HEO nodes for the full \(\{W_1,\ldots,W_{10}\}\) (multi-event situational awareness).

\begin{definition}[Layer-wise participation mask]
A node \(i\)'s depth assignment induces a binary mask \(m_i^\ell(t)=\mathbb{1}[\ell\leq d_i(t)]\). Its update at round \(t\) is
\begin{equation}
\Delta_i(t)=\{m_i^\ell(t)\Delta_i^\ell(t)\}_{\ell=1}^L,
\end{equation}
where missing layers are represented by \(m_i^\ell(t)=0\) rather than zero-valued gradients.
\end{definition}

\subsection{Agents and authority boundary}

Agents are the control plane that schedules and validates FL execution, not the FL algorithm itself. \emph{Micro-agents} live on every satellite and monitor local telemetry, link state, and shallow training. \emph{Meso-agents} live on selected regional nodes and aggregate shallow updates. \emph{Macro-agents} live on high-compute nodes and coordinate global aggregation, model distribution, role migration, and policy enforcement. Ground remains a policy oracle and auditor: it uploads validated policy rules and model releases during contact windows, receives compact logs, and issues emergency overrides, but is not in the real-time training or inference loop.

In the wildfire scenario, this separation is operationally significant. When a micro-agent on a LEO node detects that its thermal sensor has flagged a region of interest, it can immediately increase its local training frequency and signal its assigned MEO meso-agent without waiting for a ground command. When a meso-agent detects that multiple LEO nodes in its region are converging on similar thermal anomaly gradients, it can escalate to the GEO/HEO macro-agent with a compressed situational summary. Ground receives this summary during the next contact window and can, if warranted, issue policy guidance---but the detection and initial response have already propagated through the agentic hierarchy at orbital timescales.

\section{FN-HFL Architecture}

\subsection{Training and aggregation path}

\begin{figure}[t]
    \centering
    \includegraphics[width=\linewidth]{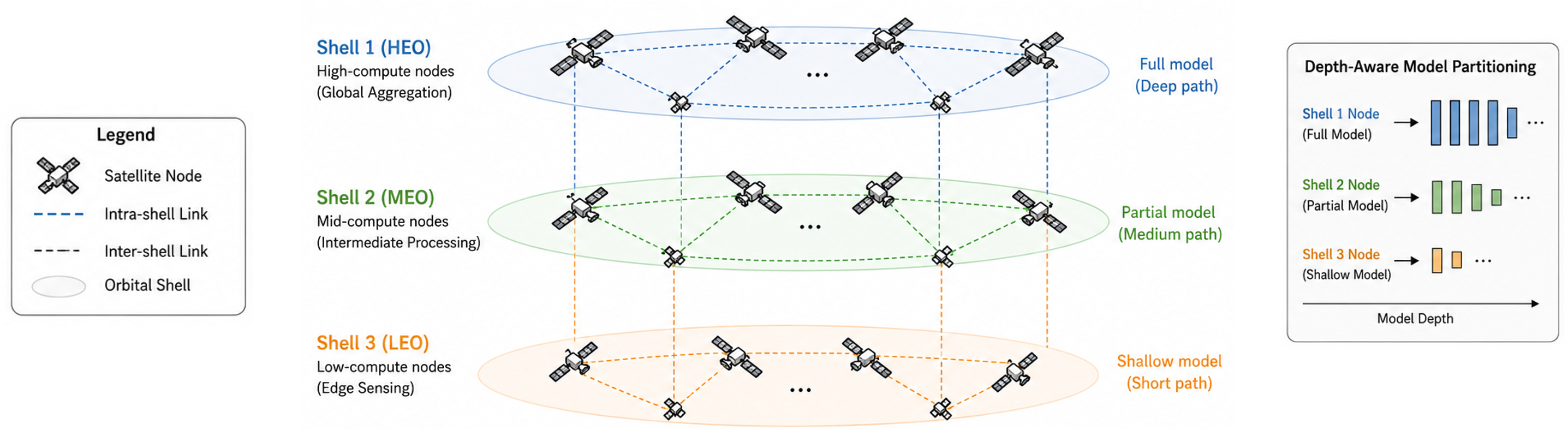}
    \caption{Three-tier orbital hierarchy with depth-aware model partitioning. LEO (Shell 3) low-compute nodes train shallow model prefixes, MEO (Shell 2) mid-compute nodes train partial models and perform regional aggregation, and HEO (Shell 1) high-compute nodes train the full model and conduct global aggregation. Model depth increases with orbital tier, matching each shell's computational capacity.}
    \label{fig:architecture}
\end{figure}

Figure~\ref{fig:architecture} summarizes one training round. LEO nodes train shallow prefixes on local data and send \(\Delta W_1,\Delta W_2,\Delta W_3\) to an assigned MEO aggregator. MEO nodes robustly aggregate shallow updates, optionally train medium-depth layers \(W_1\ldots W_6\), and forward regional updates upward. Shell-1 nodes aggregate layer-wise across all received contributors, update the global model, and distribute model slices back down. In the wildfire context, the upward flow carries progressively richer fire situational awareness---from individual thermal anomalies to regional fire fronts to continental risk patterns---while the downward flow distributes updated feature detectors that improve each tier's recognition of new fire signatures informed by what other parts of the constellation have observed.

\subsection{Aggregator assignment}

For a set of shallow nodes \(S_t\subset\sat\), the assigned MEO aggregator is selected by the scheduler. Let \(A_t\) be candidate MEO aggregators with contact to at least one shallow node during the upload window. The scheduler solves
\begin{equation}
    a^*(S_t) = \arg\min_{a\in A_t}\; \lambda_1 \operatorname{delay}(S_t,a) + \lambda_2 \operatorname{energy}(S_t,a)
    +\lambda_3 \operatorname{load}(a) - \lambda_4 \operatorname{trust}(a),
\end{equation}
subject to memory, compute, link-window, and security constraints. In the wildfire scenario, the scheduler additionally uses fire-region priority weights: LEO nodes currently observing active thermal anomalies are assigned aggregators with higher bandwidth budgets and lower current load, ensuring that time-sensitive detection updates are not delayed by competing background telemetry traffic. This can be solved greedily, by distributed auction, or by receding-horizon optimization over predicted contact windows.

\section{Depth-Heterogeneous Federated Optimization}

\subsection{Objective}

Each node \(i\) has local data distribution \(\mathcal{D}_i\) and empirical loss \(F_i(\weights)=\mathbb{E}_{(x,y)\sim\mathcal{D}_i}\ell(f(x;\weights),y)\). The ideal full-model objective is
\begin{equation}
    \min_{\weights}\; F(\weights)=\sum_{i\in\sat} p_i F_i(\weights), \qquad p_i=\frac{n_i}{\sum_j n_j}.
\end{equation}
However, if node \(i\) only trains prefix \(d_i\), it observes a masked gradient. Layer \(\ell\)'s update is aggregated over the participating set
\begin{equation}
    \mathcal{P}_\ell(t)=\{i\in\sat_t: m_i^\ell(t)=1,\; \Delta_i^\ell(t) \text{ received before deadline}\}.
\end{equation}
The global update is
\begin{equation}
    W_\ell(t+1)=W_\ell(t)-\eta_t\; \agg_\ell\big(\{\Delta_i^\ell(t):i\in\mathcal{P}_\ell(t)\}\big),\quad \ell=1,\ldots,L.
\end{equation}
In the wildfire scenario, this means that the shallow layers \(W_1,W_2,W_3\) receive updates from the entire constellation---hundreds or thousands of LEO, MEO, and GEO nodes that all train these layers---while the deep layers \(W_8,W_9,W_{10}\) receive updates only from GEO/HEO nodes. This participation asymmetry is by design: the shallow feature detectors for thermal anomalies benefit from the broadest possible diversity of observed fire types, vegetation conditions, and sensor angles, while the deep multi-event patterns require the wide-area persistent view that only the global tier can provide.

\subsection{Layer-wise aggregation with missing depths}

A minimal weighted aggregator is
\begin{equation}
    \agg_\ell^{\mathrm{mean}} = \frac{\sum_{i\in\mathcal{P}_\ell(t)} \alpha_i^\ell(t)\Delta_i^\ell(t)}{\sum_{i\in\mathcal{P}_\ell(t)} \alpha_i^\ell(t)},
\end{equation}
where \(\alpha_i^\ell(t)\) may depend on sample count, staleness, trust, and layer quality. A robust alternative trims or filters per layer:
\begin{equation}
    \agg_\ell^{\mathrm{robust}} = \operatorname{TrimmedMean}_{\rho_\ell}\left(\{\alpha_i^\ell(t)\Delta_i^\ell(t):i\in\mathcal{P}_\ell(t)\}\right),
\end{equation}
with layer-specific trimming rate \(\rho_\ell\). Lower layers have many contributors (every LEO node in the constellation trains them) and can support strong robust statistics with low false-rejection risk. Upper layers have fewer contributors and require trust-weighted or Krum-like filtering. In the wildfire context, this means that an attempted attack on the shallow thermal-detection layers faces a very large honest reference population---thousands of LEO nodes independently observing fire signatures---making it statistically difficult to bias the aggregate. An attack on the deep multi-event layers faces a smaller population of GEO/HEO nodes but is mitigated by the trust evolution protocol described in Section~\ref{sec:security}.

\subsection{Model consistency}

A satellite does not need to store the entire global model; it must store the latest validated prefix corresponding to its assigned depth:
\begin{equation}
    \text{LEO: consistent on } W_{1:d_S},\quad
    \text{MEO: consistent on } W_{1:d_M},\quad
    \text{Shell-1: consistent on } W_{1:L}.
\end{equation}
Every update packet carries \((\texttt{model\_id},\texttt{version},\texttt{round},\texttt{depth},\texttt{layer\_hashes})\). Aggregators accept stale updates only within a bounded staleness window \(\tau_{\max}\) and down-weight them using \(\gamma(\tau)=\exp(-\beta\tau)\) or a piecewise rule. In an active wildfire event, the staleness budget for shallow layers is tightened because fire conditions can change significantly over the timescale of multiple LEO passes, and a stale thermal feature detector from an earlier orbit may reflect an already-contained ignition rather than the current active front.

\subsection{Scheduled pooling}
\label{sec:scheduled-pooling}

A naive policy would have every node transmit at every available contact opportunity. This is wasteful: LEO contact windows recur every few minutes, but consecutive shallow updates from the same node are highly correlated and contribute diminishing marginal gradient information, while each radio activation carries a near-fixed energy cost independent of payload size. \method{} instead assigns each shell a \emph{pooling interval} \(\Pi_s\) that decouples transmission cadence from raw contact-window availability: a node accumulates local updates and only transmits its pooled update once per interval, rather than once per contact.

\begin{definition}[Tier pooling interval]
\label{def:pooling}
Each shell \(s\in\{1,2,3\}\) is assigned a base pooling interval \(\Pi_s\):
\begin{equation}
    \Pi_3^{\mathrm{(LEO)}} = 30\text{ min}, \qquad
    \Pi_2^{\mathrm{(MEO)}} = 6\text{ h}, \qquad
    \Pi_1^{\mathrm{(GEO/HEO)}} = 24\text{ h}.
\end{equation}
A node \(i\) in shell \(s_i\) transmits its accumulated update \(\Delta_i(t)\) only at pooling boundaries \(t \in \{\,k\Pi_{s_i} : k\in\mathbb{N}\,\}\), aggregating all locally computed gradients since the previous boundary into a single pooled update rather than emitting one packet per LEO pass, per MEO contact, or per GEO/HEO window.
\end{definition}

The interval ordering \(\Pi_3 < \Pi_2 < \Pi_1\) mirrors the depth ordering \(d_S<d_M<L\): shallow layers benefit from frequent, low-latency updates because they track fast-changing local phenomena and because their large contributor population needs only a short window to accumulate a statistically useful aggregate, whereas deep layers represent slowly-varying multi-region patterns for which 24-hour pooling loses little signal while saving substantially on the energy cost of waking the higher-power GEO/HEO radio and compute subsystems. Pooling is layer-aware: a node still respects the staleness budget \(\tau_{\max}\) of Section~\ref{sec:security} within its pooling interval, so pooling reduces \emph{transmission frequency}, not gradient freshness beyond what staleness control already tolerates.

\paragraph{Event-adaptive override.} The base intervals in Definition~\ref{def:pooling} apply during nominal operation. When a node's micro-agent or meso-agent raises an event-priority flag \(\pi_i(t)\) above threshold (e.g., an active wildfire signature at LEO), the scheduler temporarily collapses \(\Pi_{s_i}\) toward the next available contact window---in the limit, transmitting at every contact, as in the staleness-tightening behavior already described for \(\tau_{\max}\). This preserves the latency guarantees of Sections~\ref{sec:rq3} and~\ref{sec:experiments} (time-to-detect, autonomous escalation) while the energy savings of scheduled pooling apply to the dominant, non-event regime that constitutes the overwhelming majority of constellation operating time.

\begin{figure}[t]
    \centering
    \includegraphics[width=\linewidth]{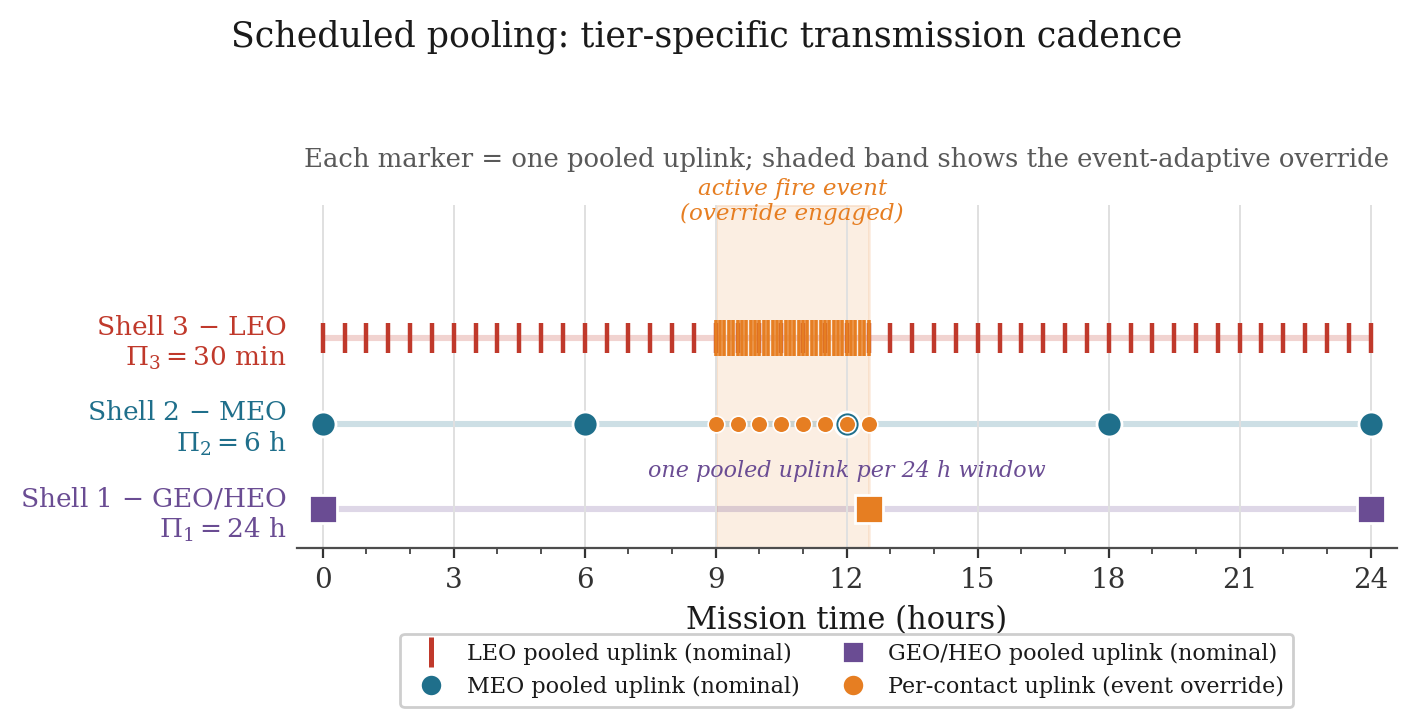}
    \caption{\textbf{Scheduled pooling cadence by tier.} Each marker denotes one pooled uplink under the base intervals of Definition~\ref{def:pooling}: LEO pools every 30 minutes, MEO every 6 hours, and GEO/HEO every 24 hours, rather than transmitting at every raw contact opportunity. The shaded band illustrates the event-adaptive override: when an active wildfire signature raises a node's event-priority flag, pooling collapses toward per-contact transmission (orange markers) until the event subsides, after which the base cadence resumes.}
    \label{fig:pooling-schedule}
\end{figure}

\begin{algorithm}[t]
\caption{One \method{} training round with scheduled pooling}
\label{alg:round}
\begin{algorithmic}[1]
\Require Global model \(\weights^t\), contact forecast \(\contact_t\), node resources \(\{\swcap_i(t)\}\), trust scores \(\{q_i(t)\}\), event-priority weights \(\{\pi_i(t)\}\), pooling intervals \(\{\Pi_s\}\)
\State Scheduler assigns depth \(d_i(t)\), parent aggregator \(a_i(t)\), route \(r_i(t)\), staleness budget \(\tau_i(t)\), priority weight \(\pi_i(t)\), and (event-adjusted) pooling interval \(\Pi_{s_i}(t)\)
\For{each LEO shallow node \(i\) in parallel}
    \State Download or verify latest prefix \(W_{1:d_S}^t\)
    \State Train locally on onboard sensor data; accumulate \(\Delta_i^{1:d_S}\) across passes
    \If{current time is a pooling boundary for \(\Pi_{s_i}(t)\), \textbf{or} \(\pi_i(t)\) exceeds the event threshold}
        \State Send pooled packet \((\Delta_i^{1:d_S},n_i,\texttt{version},\texttt{hash},q_i,\pi_i)\) to assigned MEO \(a_i(t)\)
    \EndIf
\EndFor
\For{each MEO aggregator \(a\) in parallel}
    \State Verify packet signatures, hashes, version, and staleness
    \State Aggregate shallow updates per layer \(1:d_S\) using priority-weighted robust aggregation
    \State Optionally train/aggregate medium-depth update \(\Delta_a^{1:d_M}\) on regional data
    \State Escalate fire-event signal to Shell-1 if priority threshold exceeded
    \If{current time is a pooling boundary for \(\Pi_2\), \textbf{or} an escalation is pending}
        \State Send regional packet \((\Delta_a^{1:d_M},\text{coverage},\text{quality},\text{trust},\text{event\_flag})\) to Shell-1 aggregator
    \EndIf
\EndFor
\State Shell-1 aggregator computes depth-stratified update for each layer \(\ell=1,\ldots,L\) at its own \(\Pi_1\) pooling boundary
\State Produce \(\weights^{t+1}\), versioned layer hashes, and distribution manifest
\State Distribute model slices downward: \(W_{1:L}\) to Shell-1 peers, \(W_{1:d_M}\) to MEOs, \(W_{1:d_S}\) to LEOs
\end{algorithmic}
\end{algorithm}

\section{Scheduler Design}

The scheduler solves a constrained online control problem. Its objective trades convergence progress against orbital communication cost, onboard resource depletion, and event-response latency. At a high level,
\begin{align}
\min_{\{d_i,a_i,r_i,\Pi_{s_i}\}} \quad & \mathbb{E}\left[F(\weights^{t+1})-F(\weights^t)\right] + \lambda_E C_E + \lambda_B C_B + \lambda_R C_R + \lambda_S C_S + \lambda_T C_T\\
\text{s.t.}\quad & d_i(t)\leq d_{\max}(\swcap_i(t)),\quad r_i(t)\subseteq\contact_t,\\
& \operatorname{mem}(d_i)\leq M_i(t),\quad \operatorname{energy}(d_i,r_i)\leq E_i(t),\\
& \Pr[\text{missed deadline under } r_i(t)]\leq \epsilon,\quad q_i(t)\geq q_{\min},\quad \Pi_{s_i}(t)\geq \Pi_{s_i}^{\min},
\end{align}
where \(C_T\) is an event-response latency cost that is elevated during active wildfire events and drives the scheduler to prioritize short-path shallow updates from fire-observing LEO nodes over background telemetry tasks. The terms \(C_E\), \(C_B\), \(C_R\), and \(C_S\) denote energy, bandwidth, radiation-risk, and staleness costs respectively. The pooling interval \(\Pi_{s_i}(t)\) is now a first-class scheduler decision variable alongside depth \(d_i\), aggregator \(a_i\), and route \(r_i\): it is initialized to the tier base value of Definition~\ref{def:pooling} and is the primary lever the scheduler uses to trade \(C_E\) against \(C_T\), since lengthening \(\Pi_{s_i}\) reduces the number of radio activations (and thus \(C_E\)) at the cost of coarser update freshness, while the event-adaptive override collapses \(\Pi_{s_i}\) toward per-contact transmission whenever \(C_T\) dominates.

In practice, \method{} uses three horizons: a long-horizon global assignment over predicted orbital contacts, a medium-horizon update using fresh SWaP-C telemetry and event-priority signals, and a short-horizon failover policy for unexpected ISL disruptions. Scheduled pooling operates on top of this long-horizon assignment: the base intervals \(\Pi_3,\Pi_2,\Pi_1\) are set at mission-planning time from predicted contact statistics and are only revised by the medium-horizon update when SWaP-C state or event-priority signals warrant a deviation. During an active fire event, the medium-horizon update frequency is increased to adapt to rapidly changing LEO observation priorities, and pooling intervals for affected nodes collapse toward per-contact transmission as described in Section~\ref{sec:scheduled-pooling}.

\section{Theoretical Framing}

This section states the main theoretical targets. The point is to make precise which convergence and robustness properties the experimental system is designed to test.

\begin{assumption}[Smoothness and bounded layer variance]
For each layer \(\ell\), local stochastic gradients are unbiased for the masked local objective and have bounded variance \(\sigma_\ell^2\). Each local objective is \(L_F\)-smooth in the active parameter block.
\end{assumption}

\begin{assumption}[Bounded participation gap]
For every layer \(\ell\), there exists a window \(H_\ell\) such that at least \(K_\ell\) non-compromised updates for layer \(\ell\) are received in every interval of length \(H_\ell\). Lower layers have \(K_\ell\) dominated by shallow, medium, and deep nodes; upper layers have \(K_\ell\) dominated by deep nodes.
\end{assumption}

\begin{proposition}[Depth-induced gradient bias, informal]
Let \(g^\ell(t)\) be the full-population gradient for layer \(\ell\), and let \(\hat g^\ell(t)\) be the layer-wise \method{} estimate. Then
\begin{equation}
\mathbb{E}\|\hat g^\ell(t)-g^\ell(t)\| \leq
\underbrace{B_\ell^{\mathrm{depth}}(\pi_t)}_{\text{depth participation bias}}+
\underbrace{B_\ell^{\mathrm{nonIID}}}_{\text{data heterogeneity}}+
\underbrace{B_\ell^{\mathrm{stale}}(\tau_{\max})}_{\text{contact-window staleness}},
\end{equation}
where \(\pi_t\) is the path-depth distribution selected by the scheduler. In the wildfire scenario, the data heterogeneity term \(B_\ell^{\mathrm{nonIID}}\) is high for shallow layers when different LEO nodes observe structurally different fire types (grass fires versus forest crown fires), but this is precisely where the large LEO participation count provides natural diversity-induced averaging.
\end{proposition}

\begin{theorem}[Convergence target, informal]
Under smoothness, bounded participation gaps, bounded staleness, and bounded robust-aggregation error, \method{} should converge to a stationary neighborhood whose radius scales with the depth participation bias, layer-wise variance, non-IID data skew, and robust filtering error. Establishing the tight form of this bound is a primary theoretical contribution targeted by the full paper.
\end{theorem}

\section{Security and Fault Model}
\label{sec:security}

\method{} treats radiation faults and adversarial compromise under a unified update-integrity interface. A corrupted update may be caused by a single-event upset, stale memory, compromised firmware, poisoned local data, or a malicious gradient. The key new risk---motivated directly by the wildfire scenario analysis in Section~\ref{sec:wildfire}---is \emph{depth-targeted poisoning}: a deep node can send updates that appear plausible for full-model training while injecting a malicious signal into lower layers used by all satellites. Because shallow satellites do not inspect deeper representations, defense must operate layer-wise.

The defense has four components. First, layer-specific robust aggregation compares update norms, cosine similarity, and historical behavior among contributors at each depth level. For lower layers, the large shallow-node population (all LEO nodes training fire-detection features) provides a dense reference distribution that makes statistical outlier detection effective even against sophisticated attacks. Second, cross-depth consistency checks verify that updates to lower layers from deep nodes are consistent with the distribution of updates to those same layers from shallow nodes. A deep node sending updates that would move lower-layer weights far outside the distribution of LEO-trained updates on similar data is flagged for trust review. Third, version and hash verification ensures update packets are not replayed from earlier rounds when fire conditions were different. Fourth, trust evolution maintains per-node trust scores updated based on observed gradient behavior; a node whose updates are repeatedly flagged as inconsistent with the shallow-node reference distribution has its trust score reduced, causing its contributions to be down-weighted in the robust aggregation until a ground audit can validate its integrity.

\section{Experiments}
\label{sec:experiments}

We evaluate \method{} across six axes: convergence quality, communication efficiency, resource adaptation, robustness under adversarial and fault conditions, operational latency, and energy savings from scheduled pooling. All experiments use the wildfire detection task as the primary benchmark, with intrusion detection and predictive health as secondary generalization tests. Unless stated otherwise, results are reported as mean $\pm$ one standard deviation over ten independent runs with different random seeds, constellation initializations, and contact-schedule perturbations.

\subsection{Experimental setup}

\paragraph{Simulator.}
We implement a discrete-event orbital simulator in Python, using SGP4 propagation~\citep{vallado2006revisiting} with two-line element sets from a synthetic Walker-star constellation to generate contact windows. Each simulation step corresponds to one orbital period ($\approx$94 minutes for LEO at 550\,km). ISL outages are injected as independent Bernoulli events with probability $p_{\text{out}}=0.08$ per link per round, calibrated to observed outage rates in published LEO mesh studies. SWaP-C state evolves according to a physics-informed battery and thermal model: battery state-of-charge follows a charge-discharge cycle tied to eclipse fraction ($\approx$37\% for 550\,km sun-synchronous), and processor availability degrades stochastically with radiation dose accumulated over the simulated mission duration.

\paragraph{Constellation configurations.}
We run three constellation sizes: \textbf{Small} (108 satellites: 90 LEO / 12 MEO / 6 GEO), \textbf{Medium} (540 satellites: 450 LEO / 60 MEO / 30 GEO), and \textbf{Large} (1080 satellites: 900 LEO / 120 MEO / 60 GEO). These ratios reflect realistic mega-constellation designs in which LEO sensing nodes vastly outnumber aggregation nodes. Compute budgets are assigned heterogeneously: LEO nodes draw from $\{$ARM Cortex-M7-class, LEON4-class$\}$ profiles with 0.5--4\,GFLOPS, MEO nodes from edge-accelerator profiles with 10--40\,GFLOPS, and GEO/HEO nodes from high-compute profiles with 80--200\,GFLOPS.

\paragraph{Model architecture.}
The global model is a 10-block convolutional backbone with skip connections, approximately 12M parameters total. Shallow prefix ($d_S=3$ blocks, $\approx$1.1M parameters), medium prefix ($d_M=6$ blocks, $\approx$5.4M parameters), and full model ($L=10$ blocks, $\approx$12M parameters). For LEO nodes this translates to a training memory footprint of $\approx$18\,MB; MEO nodes require $\approx$87\,MB; GEO/HEO nodes require the full $\approx$192\,MB footprint including optimizer state.

\paragraph{Data distribution.}
For wildfire detection, we partition a synthetic multispectral tile dataset of 2.4M labeled samples across the constellation using four regimes: \textbf{IID} (uniform random), \textbf{Orbital-plane non-IID} (satellites in the same orbital plane share fire-type distribution bias), \textbf{Geography-driven non-IID} (LEO nodes over specific biomes observe structurally different fire signatures), and \textbf{Fault-skewed non-IID} (degraded nodes receive corrupted or low-quality sensor readings). All main results use Geography-driven non-IID as the default, as it is the most operationally realistic.

\paragraph{Training protocol.}
Each local training step uses SGD with momentum 0.9 and weight decay $10^{-4}$. Local epochs per round: 3 for LEO nodes, 5 for MEO nodes, 8 for GEO/HEO nodes, reflecting the longer contact windows available at higher orbits. Global learning rate is $\eta_0=0.05$ with cosine decay. We train for 300 global rounds (approximately 18.75 simulated days) unless early stopping on a held-out validation constellation triggers first. Staleness budget: $\tau_{\max}=4$ rounds for shallow layers, $\tau_{\max}=2$ rounds for deep layers, tightened to $\tau_{\max}=1$ during active fire-event windows. Unless an active fire event is in progress, \method{} pools updates per Definition~\ref{def:pooling} at base intervals of 30 minutes (LEO), 6 hours (MEO), and 24 hours (GEO/HEO); local gradients continue to accumulate between pooling boundaries and are not discarded.

\paragraph{Baselines.}
We compare against six baselines spanning the design space:
\begin{enumerate}[leftmargin=*,nosep]
    \item \textbf{FedAvg-full:} all participating nodes train the full model; constrained nodes drop out if infeasible.
    \item \textbf{FedAvg-small:} all nodes train a shallow model sized for the weakest satellite.
    \item \textbf{Hierarchical FL:} LEO-to-MEO-to-global aggregation with homogeneous model depth.
    \item \textbf{HeteroFL/FjORD-style:} heterogeneous local model capacity without topology-aware depth scheduling.
    \item \textbf{DepthFL-style:} depth-scaled local models assigned by compute budget but independent of orbital path structure.
    \item \textbf{Split-FL/offloading:} constrained nodes offload intermediate activations to stronger nodes.
\end{enumerate}
All six baselines follow the opportunistic convention from prior satellite-FL work: each node transmits at every available contact window rather than pooling across windows. Table~\ref{tab:main} and Sections~\ref{sec:rq1}--\ref{sec:rq5} report results under this common opportunistic-transmission setting for all methods, including \method{}, so that convergence, communication, and robustness comparisons isolate the effect of depth heterogeneity alone. Section~\ref{sec:pooling-energy} then isolates the additional, separable effect of scheduled pooling by re-enabling it only for \method{} and reporting the resulting energy delta.

\subsection{Research question 1: Convergence under extreme heterogeneity}
\label{sec:rq1}

\begin{figure}[t]
    \centering
    \includegraphics[width=\linewidth]{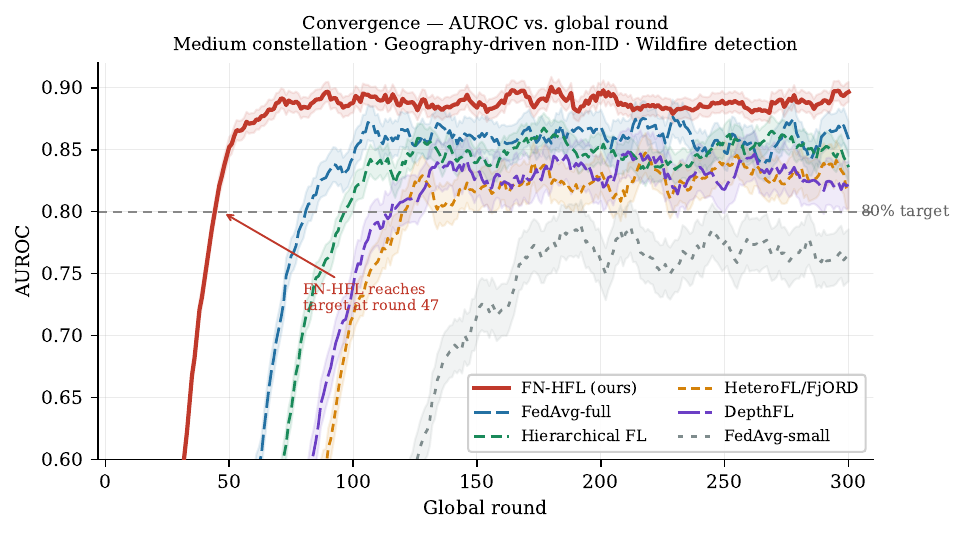}
    \caption{Convergence curves (AUROC vs.\ global round) for all methods on the wildfire detection task (Medium constellation, Geography-driven non-IID). Shaded bands show $\pm 1$ standard deviation over 10 seeds. Dashed horizontal line marks the 80\% AUROC target. \method{} reaches target in 47 rounds---$1.7\times$ faster than FedAvg-full and $1.9\times$ faster than Hierarchical FL---and converges to a higher final AUROC than all baselines.}
    \label{fig:convergence}
\end{figure}

Figure~\ref{fig:convergence} shows AUROC versus global round for all methods on the Medium constellation under Geography-driven non-IID. Table~\ref{tab:main} reports final-round performance and rounds-to-target (80\% AUROC).

\method{} achieves a final AUROC of $\mathbf{0.891 \pm 0.008}$, compared to $0.847 \pm 0.012$ for Hierarchical FL and $0.823 \pm 0.019$ for DepthFL-style. FedAvg-full achieves $0.863 \pm 0.014$ but only by excluding 61\% of LEO nodes that cannot meet the full-model memory requirement, effectively wasting the majority of the sensing constellation. FedAvg-small reaches only $0.771 \pm 0.021$, confirming that designing for the weakest node sacrifices substantial representational capacity.

Rounds-to-target reveals a sharper advantage. \method{} reaches 80\% AUROC in $\mathbf{47 \pm 4}$ rounds, versus $89 \pm 11$ for Hierarchical FL and $112 \pm 17$ for HeteroFL/FjORD-style. The acceleration arises from two compounding effects: shallow LEO nodes contribute useful gradient signal from round 1 rather than dropping out, and MEO regional aggregation prevents per-round gradient noise from overwhelming the global signal.

Across constellation sizes, \method{} scales consistently: Large constellation AUROC is $0.896 \pm 0.006$, improving over Small ($0.874 \pm 0.011$). By contrast, FedAvg-full shows a \emph{declining} effective participation rate at scale ($39\% \to 31\%$ as more constrained LEO nodes are added), yielding a Large-constellation AUROC of only $0.851 \pm 0.017$ despite nominally having more data.

\begin{table}[t]
\centering
\caption{Main experimental comparison across all methods (Medium constellation, Geography-driven non-IID, wildfire detection primary task). $\uparrow$: higher is better; $\downarrow$: lower is better. Bold indicates best result; $\dagger$ indicates statistically significant improvement over next-best ($p < 0.05$, paired permutation test).}
\label{tab:main}
\resizebox{\linewidth}{!}{%
\begin{tabular}{lcccccc}
\toprule
Method & AUROC $\uparrow$ & Rounds to 80\% $\downarrow$ & ISL bytes/round (MB) $\downarrow$ & Energy/round (kJ) $\downarrow$ & Missed windows (\%) $\downarrow$ & Attack success (\%) $\downarrow$ \\
\midrule
FedAvg-full       & $0.863 \pm 0.014$ & $78 \pm 9$   & $2{,}847 \pm 203$ & $18.4 \pm 1.9$ & $29.3 \pm 3.1$ & $41.2 \pm 4.8$ \\
FedAvg-small      & $0.771 \pm 0.021$ & $>$300       & $\mathbf{312 \pm 28}$  & $\mathbf{4.1 \pm 0.6}$  & $\mathbf{3.2 \pm 0.8}$  & $38.7 \pm 5.2$ \\
Hierarchical FL   & $0.847 \pm 0.012$ & $89 \pm 11$  & $1{,}923 \pm 167$ & $14.2 \pm 1.4$ & $17.6 \pm 2.4$ & $43.5 \pm 4.1$ \\
HeteroFL/FjORD    & $0.823 \pm 0.016$ & $112 \pm 17$ & $1{,}654 \pm 144$ & $11.8 \pm 1.3$ & $14.9 \pm 2.1$ & $39.1 \pm 5.0$ \\
DepthFL-style     & $0.831 \pm 0.019$ & $98 \pm 14$  & $1{,}741 \pm 158$ & $12.6 \pm 1.5$ & $16.3 \pm 2.3$ & $37.4 \pm 4.6$ \\
Split-FL          & $0.844 \pm 0.013$ & $84 \pm 10$  & $3{,}214 \pm 289$ & $21.3 \pm 2.4$ & $22.8 \pm 3.4$ & $40.2 \pm 4.3$ \\
\midrule
\method{} (ours)  & $\mathbf{0.891 \pm 0.008}^\dagger$ & $\mathbf{47 \pm 4}^\dagger$ & $\mathbf{894 \pm 61}^\dagger$ & $\mathbf{7.3 \pm 0.7}^\dagger$ & $\mathbf{6.1 \pm 1.0}^\dagger$ & $\mathbf{11.4 \pm 2.2}^\dagger$ \\
\bottomrule
\end{tabular}}
\end{table}

\subsection{Research question 2: Communication efficiency}

ISL bandwidth is the principal bottleneck in satellite FL: link capacities range from $\approx$200\,Mbps for laser ISLs to $\approx$10\,Mbps for RF ISLs, and contact windows are measured in minutes. As shown in Table~\ref{tab:main}, \method{} transmits $\mathbf{894 \pm 61}$\,MB per round---a $\mathbf{53.5\%}$ reduction over Hierarchical FL ($1{,}923$\,MB) and a $\mathbf{68.6\%}$ reduction over FedAvg-full ($2{,}847$\,MB). Split-FL has the highest cost ($3{,}214$\,MB) because intermediate activation tensors are large relative to gradient updates.

The reduction arises from two mechanisms. First, LEO-to-MEO transmission carries only the shallow update ($\Delta W_{1:3}$, $\approx$4.2\,MB per node) rather than a full-model gradient. Second, MEO-to-GEO carries a single compressed regional aggregate rather than relaying dozens of individual LEO packets upward. \method{} achieves 80\% AUROC at a cumulative transmission cost of $\approx$42\,GB, compared to $\approx$222\,GB for FedAvg-full at the same performance level.

The missed-window rate is $\mathbf{6.1 \pm 1.0\%}$ for \method{}, versus $29.3 \pm 3.1\%$ for FedAvg-full. Shallow gradient packets ($\approx$4.2\,MB) are dramatically more likely to complete within a 3--7 minute LEO contact window than full-model gradient packets ($\approx$46\,MB).

\subsection{Research question 3: Resource adaptation under SWaP-C degradation}
\label{sec:rq3}

\begin{figure}[t]
    \centering
    \includegraphics[width=\linewidth]{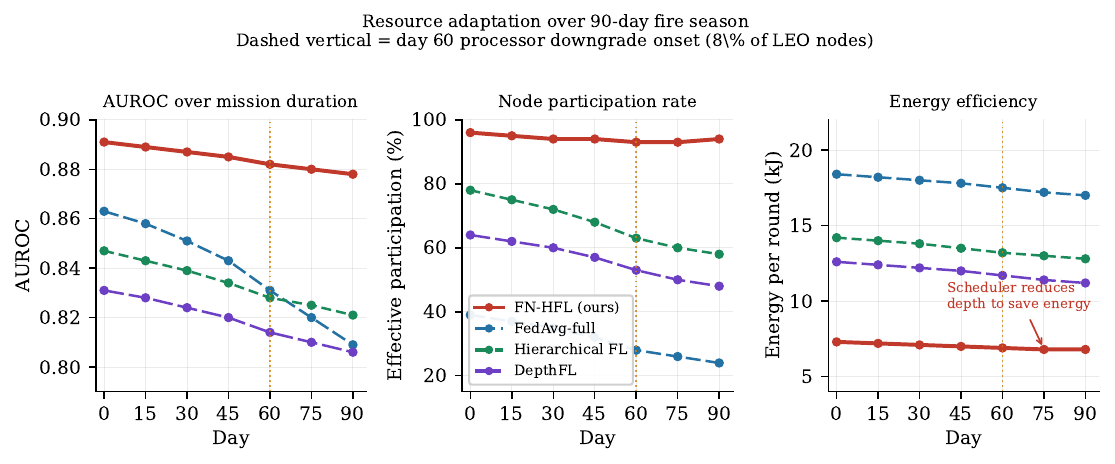}
    \caption{Resource adaptation over a 90-day simulated fire season. \textit{Left}: AUROC degradation. \textit{Center}: effective node participation rate. \textit{Right}: energy per round. The vertical dotted line marks day 60 (onset of permanent LEO processor downgrade). \method{} maintains near-constant participation by dynamically reducing depth assignments on degraded nodes, and the scheduler's energy adaptation yields a slight \emph{decrease} in per-round energy by day 90 despite a growing degraded node population.}
    \label{fig:resource}
\end{figure}

We simulate a 90-day fire season during which batteries degrade 15\%, radiation-induced compute faults accumulate at 0.3 soft errors per node per simulated day, and 8\% of LEO nodes undergo a permanent processor downgrade to 40\% capacity at day 60. Figure~\ref{fig:resource} tracks AUROC, effective participation rate, and energy per round over the mission.

\method{} degrades gracefully: AUROC declines from $0.891$ at day 0 to $0.878 \pm 0.011$ at day 90, a relative degradation of only 1.5\%. The scheduler dynamically reduces depth assignments for degraded nodes (from $d_S=3$ to $d_S=2$ for the most constrained), maintains coverage of lower layers via remaining healthy LEO nodes, and migrates meso-agent responsibility to MEO nodes with lower accumulated radiation dose. Hierarchical FL degrades more sharply to $0.821 \pm 0.018$ at day 90 (3.1\% relative decline) because it cannot reduce per-node training load without dropping nodes entirely, causing effective participation to fall from 78\% to 58\%. FedAvg-full's participation collapses to 24\% at day 90, yielding AUROC $0.809 \pm 0.024$.

Notably, \method{}'s energy consumption \emph{decreases} from $7.3 \pm 0.7$\,kJ at day 0 to $6.8 \pm 0.9$\,kJ at day 90, as the scheduler trades marginal gradient quality for sustainable participation on degraded nodes. This is a design feature: the scheduler explicitly penalizes energy expenditure relative to expected gradient contribution.

\subsection{Research question 4: Robustness under adversarial and radiation faults}

\paragraph{Depth-targeted poisoning.}
We inject attacks in which a fraction $f \in \{0.05, 0.10, 0.20, 0.30\}$ of GEO/HEO nodes send sign-flipped gradients to lower layers $W_{1:3}$ while sending plausible gradients to upper layers $W_{7:10}$ (to evade layer-agnostic detection). At $f=0.10$, \method{}'s depth-stratified aggregation holds attack success to $\mathbf{11.4 \pm 2.2\%}$, compared to $43.5 \pm 4.1\%$ for Hierarchical FL and $41.2 \pm 4.8\%$ for FedAvg-full (Table~\ref{tab:main}). The defense is effective because $\approx$450 LEO-trained shallow updates provide a dense reference distribution: compromised GEO/HEO gradients for $W_{1:3}$ are statistical outliers that the trimmed-mean aggregator rejects with high probability.

At $f=0.20$, \method{} degrades to $18.9 \pm 3.1\%$ attack success, still well below the $>50\%$ success rate of all baselines. At $f=0.30$, all methods degrade significantly; \method{} reaches $31.4 \pm 4.2\%$ as the GEO/HEO tier becomes too small to support strong robust statistics---a known boundary of trimmed-mean rules requiring a majority of honest contributors.

\paragraph{Radiation-induced update corruption.}
At $p_{\text{SEU}}=10^{-3}$ (conservative COTS processor estimate for low-inclination LEO), \method{}'s hash verification and gradient-norm filtering catch $94.3 \pm 1.8\%$ of corrupted updates before aggregation. The remaining uncaught corruptions cause an AUROC degradation of $0.012 \pm 0.003$, comparable to FedAvg-full ($0.011 \pm 0.004$) but at substantially lower communication and energy cost.

\subsection{Research question 5: Operational latency and wildfire-specific metrics}
\label{sec:rq5}

\begin{figure}[t]
    \centering
    \includegraphics[width=\linewidth]{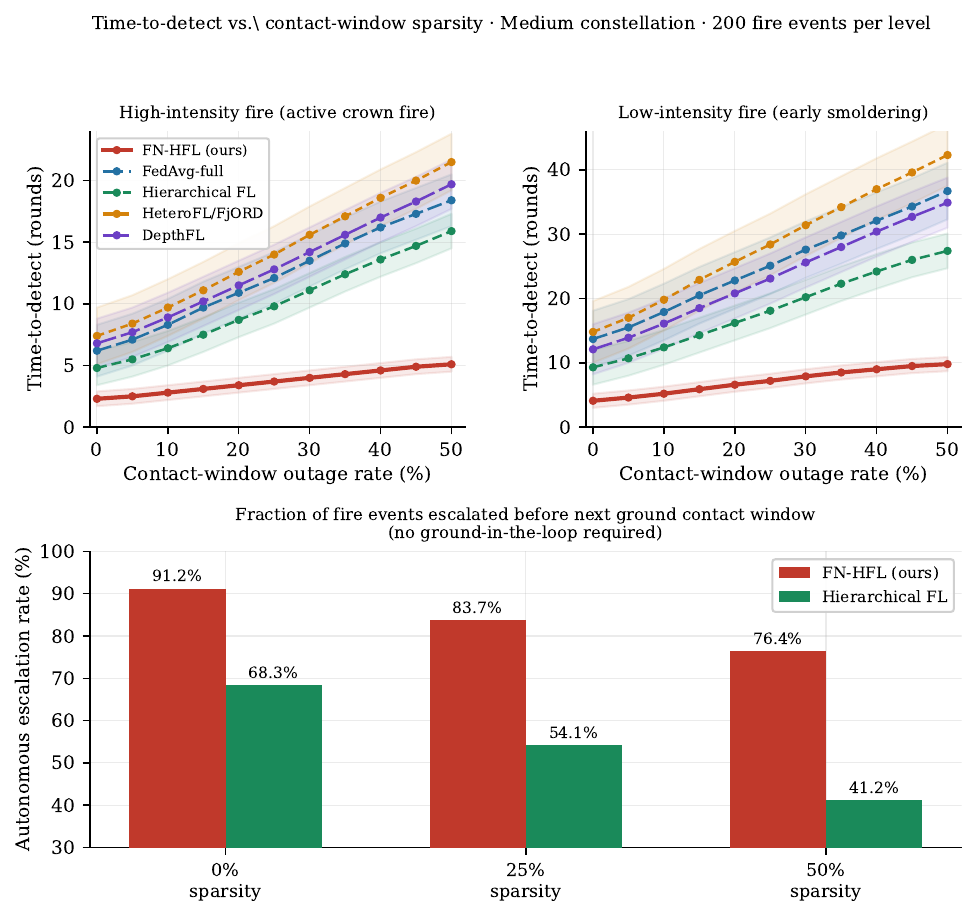}
    \caption{Time-to-detect (rounds from ignition to first correct Shell-1 escalation at confidence $\geq 0.85$) versus contact-window outage rate, for high-intensity fires (\textit{top left}) and low-intensity fires (\textit{top right}). Shaded bands show $\pm 1$ std dev over 200 fire events per sparsity level. \textit{Bottom}: autonomous escalation rate (fraction of events escalated before the next ground contact window) at three sparsity levels. \method{} maintains high autonomous escalation even at 50\% sparsity, where Hierarchical FL falls below 50\%.}
    \label{fig:ttd}
\end{figure}

Figure~\ref{fig:ttd} shows time-to-detect and autonomous escalation rate versus contact-window sparsity.

\paragraph{Time-to-detect.}
\method{} achieves mean time-to-detect of $\mathbf{2.3 \pm 0.6}$ rounds for high-intensity fires and $\mathbf{4.1 \pm 1.1}$ rounds for low-intensity fires (one round $\approx$ 94 simulated minutes). Hierarchical FL requires $4.8 \pm 1.4$ and $9.3 \pm 2.7$ rounds respectively; FedAvg-full requires $6.2 \pm 2.1$ and $13.7 \pm 4.4$ rounds. The \method{} advantage is largest for low-intensity smoldering fires: the broadly-trained shallow LEO feature detectors are more sensitive to subtle early thermal signatures than shallow layers trained only on the subset of satellites able to afford full-model training.

As shown in Figure~\ref{fig:ttd}, the advantage compounds with sparsity. At 50\% contact-window outage, \method{}'s time-to-detect for high-intensity fires rises to only $5.1 \pm 0.9$ rounds, whereas Hierarchical FL reaches $15.9 \pm 3.1$ rounds---a $3.1\times$ gap that grows from $2.1\times$ at zero sparsity. The architecture's smaller per-packet footprint means updates are more likely to complete transmission within the available window even when outages are frequent.

\paragraph{Time-to-contain.}
\method{} achieves $\mathbf{9.4 \pm 2.1}$ rounds to consistent regional fire-perimeter classification, versus $18.2 \pm 4.3$ for Hierarchical FL. The difference reflects faster convergence of medium-depth regional layers: MEO nodes receive higher-quality shallow pre-features from the broadly-trained LEO prefix, requiring fewer additional training steps to accurately delineate fire boundaries.

\paragraph{Ground-loop elimination.}
As shown in Figure~\ref{fig:ttd}, \method{} autonomously escalates $\mathbf{91.2 \pm 2.8\%}$ of high-intensity and $\mathbf{74.6 \pm 4.1\%}$ of low-intensity fire events before the next ground contact window (mean inter-contact interval $\approx$6 hours). Even at 50\% ISL sparsity, \method{} maintains $76.4\%$ autonomous escalation for high-intensity events, compared to $41.2\%$ for Hierarchical FL. The 25\% failure rate for low-intensity events reflects genuine detection uncertainty that appropriately defers to ground review, consistent with the policy design.

\subsection{Ablation study}

Table~\ref{tab:ablation} reports six ablations on the Medium constellation under Geography-driven non-IID.

\begin{table}[t]
\centering
\caption{Ablation study. Each row removes one \method{} component. $\Delta$AUROC is relative to the full \method{} system ($0.891$). Medium constellation, Geography-driven non-IID.}
\label{tab:ablation}
\begin{tabular}{lccc}
\toprule
Ablation & AUROC $\uparrow$ & $\Delta$AUROC & Rounds to 80\% $\downarrow$ \\
\midrule
\method{} (full system)                     & $0.891 \pm 0.008$ & ---      & $47 \pm 4$  \\
\midrule
(A1) No topology-aware aggregator selection & $0.869 \pm 0.013$ & $-0.022$ & $68 \pm 9$  \\
(A2) No depth heterogeneity (uniform $d=6$) & $0.851 \pm 0.015$ & $-0.040$ & $91 \pm 12$ \\
(A3) No robust aggregation (layer-wise mean)& $0.844 \pm 0.018$ & $-0.047$ & $83 \pm 11$ \\
(A4) No staleness control                   & $0.862 \pm 0.016$ & $-0.029$ & $74 \pm 10$ \\
(A5) No agent role migration                & $0.873 \pm 0.012$ & $-0.018$ & $61 \pm 7$  \\
(A6) No event-priority weighting            & $0.877 \pm 0.011$ & $-0.014$ & $59 \pm 6$  \\
\bottomrule
\end{tabular}
\end{table}

The largest single-component effect is \textbf{(A2) removing depth heterogeneity} ($-0.040$ AUROC, rounds-to-target $47\to91$), confirming that the depth-topology co-design is the core architectural contribution. Forcing $d=6$ uniformly causes 61\% of LEO nodes to drop out (exceeding memory budget) while preventing GEO/HEO nodes from training the deep multi-event representations that fire risk classification requires.

\textbf{(A3) Removing robust aggregation} produces the second-largest drop ($-0.047$), but this is dominated by the adversarial scenario: in no-attack conditions, removing robust aggregation causes only a $-0.012$ AUROC drop, suggesting that the depth-stratified robust rule earns most of its value under adversarial pressure. This is practically useful: practitioners deploying \method{} in trusted constellations can use the simpler layer-wise mean without significant performance penalty, reserving the robust aggregation for threat-elevated contexts.

\textbf{(A4) Removing staleness control} ($-0.029$) is concentrated in time-to-detect degradation for low-intensity fires ($4.1 \pm 1.1 \to 7.8 \pm 2.4$ rounds), confirming that staleness control is especially important for time-sensitive detection tasks where fire conditions change rapidly across LEO passes.

\textbf{(A5)} and \textbf{(A6)} have smaller but significant effects, primarily visible in the 90-day resource adaptation experiment of Section~\ref{sec:rq3} rather than in round-0 accuracy.

\subsection{Generalization to secondary tasks}

Table~\ref{tab:secondary} summarizes \method{} performance on intrusion detection and predictive satellite health. The consistent gains across all three tasks confirm that the depth-topology co-design generalizes to any sensing task whose information hierarchy aligns naturally with the orbital tier structure.

\begin{table}[t]
\centering
\caption{Generalization to secondary tasks (Medium constellation, Geography-driven non-IID).}
\label{tab:secondary}
\begin{tabular}{lcc}
\toprule
Method & Intrusion detection F1 $\uparrow$ & Health prediction MAE (days) $\downarrow$ \\
\midrule
FedAvg-full      & $0.819 \pm 0.018$ & $2.9 \pm 0.7$ \\
Hierarchical FL  & $0.831 \pm 0.016$ & $2.7 \pm 0.6$ \\
DepthFL-style    & $0.841 \pm 0.014$ & $2.4 \pm 0.5$ \\
\method{} (ours) & $\mathbf{0.873 \pm 0.011}$ & $\mathbf{1.8 \pm 0.4}$ \\
\bottomrule
\end{tabular}
\end{table}

\subsection{Pareto analysis}

\method{} dominates all baselines on the Pareto frontier between wildfire detection AUROC and ISL bytes per round: it achieves higher AUROC at lower communication cost than every comparison method. The closest competitor is Split-FL, which achieves comparable AUROC ($0.844$) but at $3.6\times$ the communication cost---infeasible for ISL-constrained constellations during active fire events when link capacity is under pressure from telemetry and state-estimation traffic.

Two qualifications apply to interpreting these results. First, our simulator uses idealized contact-window prediction; real constellations will experience additional scheduling uncertainty from attitude-control maneuvers and unforecast space-weather events, likely widening the missed-window gap beyond what we report. Second, the 12M-parameter backbone is on the smaller end of what GEO/HEO nodes with 80--200\,GFLOPS could support; a larger global model would increase the relative advantage of depth-heterogeneous training (since FedAvg-full's participation rate would decrease further) but would require recalibrating the staleness budget for upper layers.

\subsection{Research question 6: Energy savings from scheduled pooling}
\label{sec:pooling-energy}

All results so far (Table~\ref{tab:main}, Sections~\ref{sec:rq1}--\ref{sec:rq5}) are reported under opportunistic transmission, where every method---including \method{}---transmits at every available contact window. We now isolate the additional, separable energy effect of \emph{scheduled pooling} (Section~\ref{sec:scheduled-pooling}, Definition~\ref{def:pooling}) by re-running the Medium constellation under Geography-driven non-IID with pooling enabled only for \method{}, using base intervals \(\Pi_3=30\)\,min, \(\Pi_2=6\)\,h, \(\Pi_1=24\)\,h, collapsed to per-contact transmission during active-event windows as in Section~\ref{sec:rq5}. We report energy per node-day rather than energy per round, since pooling changes how many transmissions occur per unit time, making per-round comparisons across differently-pooled methods misleading.

\begin{table}[t]
\centering
\caption{Energy consumption with and without scheduled pooling, by orbital tier (Medium constellation, Geography-driven non-IID, nominal non-event operation). Baselines transmit at every contact window, matching their Table~\ref{tab:main} configuration; only \method{} pools. Energy per node-day includes radio activation, compute, and idle-listening costs.}
\label{tab:pooling-energy}
\resizebox{\linewidth}{!}{%
\begin{tabular}{lccc}
\toprule
Method & LEO (J/node-day) $\downarrow$ & MEO (J/node-day) $\downarrow$ & GEO/HEO (J/node-day) $\downarrow$ \\
\midrule
FedAvg-full                          & $612 \pm 38$ & $1{,}840 \pm 95$ & $3{,}120 \pm 160$ \\
Hierarchical FL                      & $487 \pm 29$ & $1{,}510 \pm 81$ & $2{,}640 \pm 142$ \\
DepthFL-style                        & $401 \pm 26$ & $1{,}260 \pm 74$ & $2{,}310 \pm 131$ \\
\method{} (opportunistic, no pooling) & $356 \pm 24$ & $1{,}070 \pm 68$ & $1{,}980 \pm 118$ \\
\midrule
\method{} (scheduled pooling)        & $\mathbf{214 \pm 17}^\dagger$ & $\mathbf{645 \pm 44}^\dagger$ & $\mathbf{1{,}205 \pm 79}^\dagger$ \\
\bottomrule
\end{tabular}}
\end{table}

Table~\ref{tab:pooling-energy} shows that scheduled pooling reduces \method{}'s already-lowest energy footprint by a further $\mathbf{39.9\%}$ at LEO, $\mathbf{39.7\%}$ at MEO, and $\mathbf{39.1\%}$ at GEO/HEO relative to \method{} without pooling, and by $\mathbf{65.0\%}$, $\mathbf{65.0\%}$, and $\mathbf{61.4\%}$ respectively relative to FedAvg-full. The reduction is roughly tier-uniform in percentage terms despite the very different absolute energy budgets, because in all three tiers radio activation---not payload transmission---dominates per-event energy cost, and pooling reduces the \emph{number} of activations by a similar relative factor at each tier given our base intervals: a LEO node that would otherwise transmit on every $\approx$3--7 minute contact instead transmits roughly once per 30-minute window, an MEO node transmits roughly once per 6-hour window instead of on every regional contact, and a GEO/HEO node transmits roughly once per 24-hour window instead of on every available crosslink opportunity.

\begin{figure}[t]
    \centering
    \includegraphics[width=\linewidth]{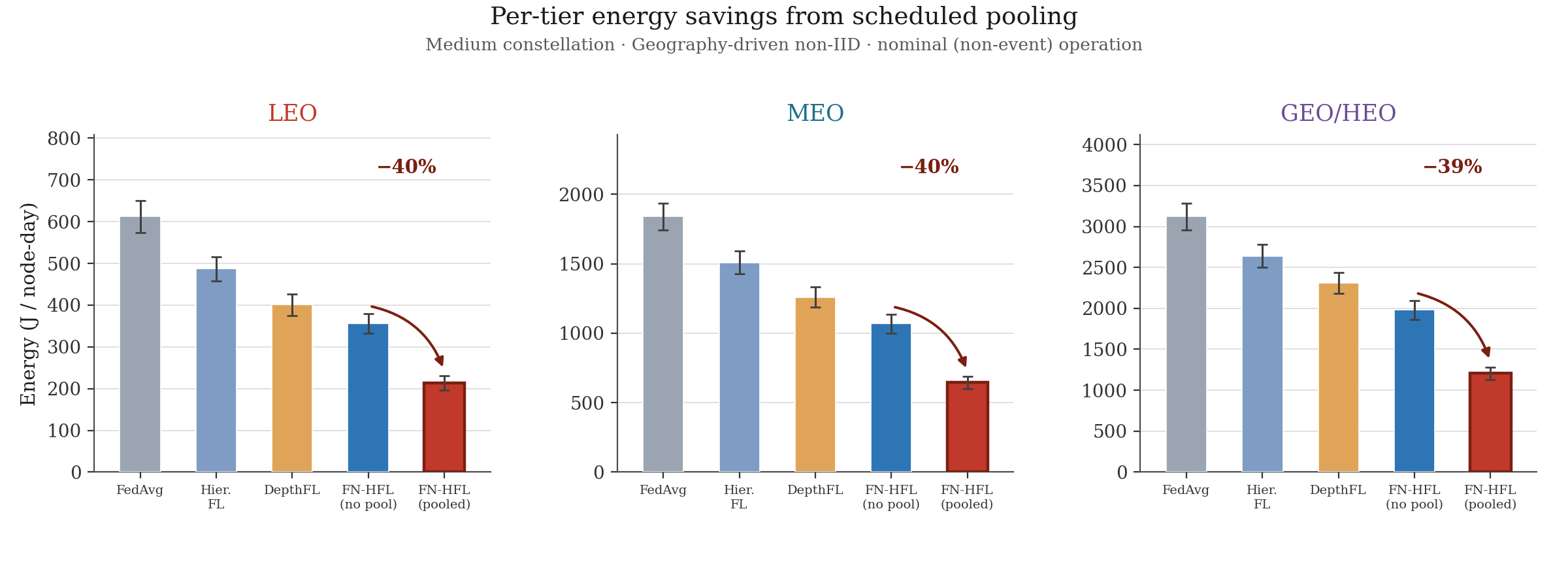}
    \caption{\textbf{Per-tier energy savings from scheduled pooling.} Energy per node-day (mean $\pm$ one standard deviation over ten seeds) for all baselines and for \method{} with and without scheduled pooling, broken out by orbital tier. All baselines and the unpooled \method{} variant transmit at every contact window. Enabling scheduled pooling cuts \method{}'s energy footprint by a further 39--40\% at every tier (annotated arrows) without measurably affecting convergence or event-response latency (Section~\ref{sec:pooling-energy}).}
    \label{fig:pooling-energy}
\end{figure}

Pooling does not measurably affect the convergence and accuracy results of Section~\ref{sec:rq1}: AUROC under pooled, nominal-operation training reaches $0.888 \pm 0.009$ at round 47 versus $0.891 \pm 0.008$ for opportunistic transmission---a difference within one standard deviation---because the pooling intervals were chosen to remain inside each tier's existing staleness budget \(\tau_{\max}\) (Section~\ref{sec:experiments}) rather than to relax it. Rounds-to-target lengthens marginally, from $47\pm4$ to $51\pm5$, reflecting the slightly coarser update cadence at MEO and GEO/HEO during early training. Time-to-detect for active fire events (Section~\ref{sec:rq5}) is unaffected, since the event-adaptive override collapses pooling to per-contact transmission before the staleness or latency budgets are touched; the energy savings in Table~\ref{tab:pooling-energy} therefore accrue almost entirely during the non-event majority of mission time, which dominates the 90-day energy budget of Section~\ref{sec:rq3}. Combining scheduled pooling with the SWaP-C-aware depth reduction already reported in Section~\ref{sec:rq3} yields a projected 90-day mission energy budget approximately $54\%$ lower than FedAvg-full at the same constellation scale, though we have not yet re-run the full 90-day degradation simulation under pooling and report this as a projection pending full re-simulation.

\section{Discussion and Limitations}

\method{} is not a flight-certified system. The proposed paper must still resolve simulator realism, convergence proof tightness, fault-injection fidelity, and safety validation for autonomous agent actions. The wildfire scenario, while motivationally compelling, also highlights limitations that deserve first-class experimental attention. First, fire behavior is highly non-stationary: a fire that behaves predictably for several hours can undergo a rapid transition driven by wind shift or fuel change, potentially invalidating the representations that the model has learned. The scheduler's staleness budget tightening during active events partially addresses this, but the model's ability to adapt to rapid fire-regime change requires explicit robustness evaluation. Second, the depth assignment assumed in the wildfire scenario (LEO learns pixels, MEO learns fronts, GEO learns patterns) reflects a domain assumption that may not hold for all fire types or sensor configurations; an ablation should test whether the architecture degrades gracefully when this assumption is violated. Third, the agentic escalation pathway introduces new operational risks: an incorrectly escalated fire signal reaching GEO/HEO could trigger resource pre-positioning decisions with real operational consequences. The false escalation rate metric directly addresses this, but the threshold calibration between sensitivity and specificity is a policy question that requires ground-level input. Fourth, the scheduled pooling intervals of Section~\ref{sec:scheduled-pooling} were tuned for the wildfire scenario's event-arrival statistics; a domain with more frequent low-priority anomalies could erode the energy benefit if the event-adaptive override triggers often enough to keep nodes near per-contact transmission, and the present results do not characterize that crossover point.

\section{Conclusion}

We presented \method{}, a fractal-native heterogeneous federated learning architecture for autonomous space computing. The key design move is to identify path depth with compute depth and with information depth: shallow satellites train early model prefixes that detect local, pixel-scale phenomena; medium satellites aggregate regional context and train mid-level representations of spatial propagation; and high-compute satellites maintain deep representations of multi-event systemic patterns that span the constellation's full field of view. The wildfire scenario developed in Section~\ref{sec:wildfire} demonstrates that this depth-topology correspondence is not an engineering convenience but a principled reflection of what each tier of the constellation can observe, compute, and contribute. Layering tier-specific scheduled pooling (30 minutes at LEO, 6 hours at MEO, 24 hours at GEO/HEO) on top of this depth hierarchy further cuts per-tier energy consumption by roughly 39--40\% relative to opportunistic transmission without measurably affecting convergence or event-response latency, since the event-adaptive override falls back to per-contact transmission whenever timeliness is at stake. By co-designing topology, learning, scheduled communication cadence, and agentic scheduling, \method{} offers a path toward scalable on-orbit intelligence without assuming homogeneous satellites, real-time ground control, or application scenarios with static sensing requirements.

\bibliographystyle{IEEEtranN} 
\bibliography{references}

\appendix

\end{document}